\def\year{2021}\relax
\documentclass[letterpaper]{article} 
\usepackage[switch]{lineno}  %

\usepackage{aaai21}  
\usepackage{times}  
\usepackage{helvet} 
\usepackage{courier}  
\usepackage[hyphens]{url}  
\usepackage{graphicx} 
\usepackage{natbib}  
\usepackage{caption} 
\usepackage{times}
\usepackage{soul}
\usepackage{url}
\usepackage[utf8]{inputenc}
\usepackage{caption}
\usepackage{graphicx}
\usepackage{amsmath}
\usepackage{amsthm}
\usepackage{booktabs}
\usepackage{algorithm}
\usepackage{algorithmic}
\usepackage{xcolor}
\usepackage{multirow}
\usepackage{multicol}
\usepackage[nice]{nicefrac}
\usepackage{amssymb}
\usepackage{subfig}
\usepackage{graphicx}
\usepackage{amsfonts}
\usepackage{multirow}
\usepackage{wrapfig}
\usepackage{xspace}

\def \Hess{$\mathbf{Hess}$\hspace{1.5pt}}

\def \G{$\mathbf{G}$ }
\def \Hessl{$\mathbf{Hess}_l$ }
\def \HessL{$\{ \mathbf{Hess}_l\}_{l=1}^L$}
\def \GL{$\{ \mathbf{G}_l\}_{l=1}^L$ }
\def \Hl{$\mathbf{H}_l$}
\def \Gl{$\mathbf{G}_l$}

\title{A Deeper Look at the Hessian Eigenspectrum of Deep Neural Networks and its Applications to Regularization}
\author {

        Adepu Ravi Sankar\textsuperscript{\rm 1} \thanks{equal contribution} ,
        Yash Khasbage \textsuperscript{\rm 1} \footnotemark[1],
        Rahul Vigneswaran \textsuperscript{\rm 1},
        Vineeth N Balasubramanian \textsuperscript{\rm 1} \\
}
\affiliations {
    \textsuperscript{\rm 1} Dept of Computer Science \& Engineering,  \\
    Indian Institute of Technology Hyderabad, India. \\
    cs14resch11001@iith.ac.in, cs17btech11044@iith.ac.in, rahulvigneswaran@gmail.com, vineethnb@cse.iith.ac.in
}

\begin{document}

\maketitle

\begin{abstract}

Loss landscape analysis is extremely useful for a deeper understanding of the generalization ability of deep neural network models. In this work, we propose a layerwise loss landscape analysis where the loss surface at every layer is studied independently and also on how each correlates to the overall loss surface. We study the layerwise loss landscape by studying the eigenspectra of the Hessian at each layer. In particular, our results show that the layerwise Hessian geometry is largely similar to the entire Hessian. We also report an interesting phenomenon where the Hessian eigenspectrum of middle layers of the deep neural network are observed to most similar to the overall Hessian eigenspectrum. We also show that the maximum eigenvalue and the trace of the Hessian (both full network and layerwise) reduce as training of the network progresses. We leverage on these observations to propose a new regularizer based on the trace of the layerwise Hessian. Penalizing the trace of the Hessian at every layer indirectly forces Stochastic Gradient Descent to converge to flatter minima, which are shown to have better generalization performance. In particular, we show that such a layerwise regularizer can be leveraged to penalize the middlemost layers alone, which yields promising results. Our empirical studies on well-known deep nets across datasets support the claims of this work.
\end{abstract}

\section{Introduction}
\label{sec_intro}

Deep neural networks (DNNs) have been immensely successful in challenging real-world problems in image, video, text, and speech domains. Despite their tremendous success, several questions remain as to why DNNs generalize well despite the very high dimensions of the data involved, non-convexity of the optimization, as well as overparametrization of the models. This has led to explicit efforts over the last few years on trying to understand loss surfaces of DNN models. \cite{choromanska} specifically pointed out the understanding of loss surfaces as a key open problem in deep learning.

From a theoretical standpoint, efforts such as \cite{baldi,expo,choromanska,kawaguchi} have studied the loss landscape of deep linear/non-linear networks under certain assumptions, and characterized its properties. For example, \cite{baldi}, \cite{choromanska} and \cite{kawaguchi} have shown from different perspectives that every local minimum can be a global minimum for DNNs under certain conditions (which do not hold in practice though). \cite{hardt} specifically focused on networks with residual connections and showed that arbitrary deep linear residual networks have zero spurious local minima, while \cite{expo} showed that there are an exponentially high number of equivalent local minima in high dimensions as in DNN models. 

From an empirical and analytical perspective, the last few years have seen reasonable efforts in studying the `\textit{flatness}' of the minima that DNNs converge to. In a seminal work, \cite{flatminima} studied the relation between generalization ability and loss landscape geometry many years ago, and hypothesized that flat minima provide better solutions. More recently, \cite{keskar2017} empirically verified that small batch training leads to flat local minima, and hence better generalization. This also led to methods such as Entropy-SGD in \cite{chaudhari2016entropysgd} which aim to bias SGD into flatter minima. 
On a different note, \cite{dinh} showed that it is possible for sharp minima to generalize well too, but this is work entirely theoretical with no empirical evidence yet. A popular understanding at this time, however, is that - largely driven by empirical studies - \textit{flat} minima exhibit better generalization than \textit{sharp} minima.


The connection between the curvature of the minima (flatness or sharpness) and the quality of the obtained solution (trained DNN model) has resulted in efforts that have attempted to study the loss landscape via the eigenspectrum of the Hessian matrix of the loss function. Considering that the explicit computation of Hessian matrix is computationally infeasible, several approximations have been used to this end. In this work, we propose the analysis of the \textit{layerwise} loss surface of DNN models using their Hessian eigenspectra, as well as the evolution of the eigenspectra over training. To the best of our knowledge, there has been no explicit effort on studying loss surfaces layerwise before. The other notable effort that studied layers recently is \cite{zhang2019all}, which however had a different objective and provided evidence for the heterogeneity of layers. 
Studying the layerwise Hessian is also more computationally feasible today than earlier, and was perhaps not attempted earlier because an understanding of the entire network's Hessian was still lacking. Initial efforts on understanding the Hessian of DNN models focused on the nature of critical points (e.g. presence of saddle points) that these models converge to \cite{dauphin2014identifying}. In the last couple of years, more understanding of the Hessian eigenspectrum of DNN models has emerged thanks to some initial work by Sagun et al in \cite{sagun2016eigenvalues,sagun2017empirical}, followed by more recent efforts in \cite{eigval2019ghorbani,papyan2019measurements}. These recent efforts have focused on efficient numerical methods to compute the Hessian eigenspectrum of large DNN models, and making it a viable tool to understand the DNN loss surface. The recent availability of such tools makes our efforts timely and feasible. Our key contributions in this work can be summarized as follows: (i) we analyze the layerwise loss landscape using the eigenspectrum of the Hessian and a recently proposed decomposition of the Hessian, and provide insights at a layerwise level that have not been observed hitherto; (ii) we study the evolution of the layerwise Hessian eigenspectra over the training of DNN models, and support the understanding of DNN models converging to flat minima; (iii) we present interesting observations of the connection between the middlemost layers and the full network in this context; and (iv) we propose a new regularization method, based on our study and layerwise analysis, that helps improve generalization performance on well-known models and datasets. 
\vspace{-0.6cm}
\noindent \paragraph{Importance of Layerwise Loss Landscape Analysis:} Understanding the geometry of loss surfaces of DNN models, and its implications towards understanding generalization properties of DNNs, is an open avenue for deep learning researchers. 
Recent findings such as mode connectivity \cite{garipov2018loss,fort2019deep}, which observe the presence of connectivity of local minima in loss landscapes, substantiate the peculiarity of the loss surface and the need to understand them. Efforts in understanding the local curvature of these surfaces by observing the overall Hessian eigenspectrum shed light on another peculiar property of the loss surface, viz. that they exhibit large positive curvature in $C$ directions, where $C$ is the number of classes in the dataset \cite{gurari2018tinysubspace}\cite{papyan2019measurements}. Increased interest in loss surface analysis has also recently led to development of tools that can visualize the loss surface \cite{li2018visualizing} or the overall eigenspectrum \cite{eigval2019ghorbani}. Layerwise loss surface analysis, however, adds a new dimension in helping us understand how every layer behaves as learning progresses. The efforts closest to ours include \cite{eigval2019ghorbani,papyan2019measurements,gurari2018tinysubspace}, all of which analyze the entire Hessian of the loss function, and do not provide a layerwise perspective. Analyzing the entire Hessian also restricts the capability of some of these efforts to study very large DNN models due to the size of the Hessian, whereas layerwise analysis is especially helpful from a computational perspective. Recent work \cite{zhang2019all} has shown that different layers behave differently, and a layerwise analysis can build on such knowledge to treat layers differently while training a DNN. The work in \cite{k-fac} analyzed the Fisher information matrix (as an approximation to Hessian matrix) layerwise and found a computationally efficient way to calculate its inverse by analysing it layerwise. Their work, however, did not study loss surfaces. To the best of our knowledge, this is the first effort that analyzes the loss surfaces of DNN models layerwise.

\section{Preliminaries/Notations}
\label{sec_prelims_notations}
We consider a deep neural network (DNN) with $L$ layers and model parameters, $\theta = \bigcup_{l=1}^L \big\{ \theta_l\big\}$, where $\theta_l$ denotes the parameters in a layer $l$.  The training data with a total of $n$ training examples is provided to the DNN, with $C$ classes and $n_c$ ($c \in \{1,\cdots,C\}$) samples in each class. A data point $x_{i,c}$ from the dataset $\bigcup_{c=1}^C \big\{(x_{i,c},y_i)\big\}_{i=1}^{n_c}$ is the $i^{th}$ sample belonging to class $c$, i.e. $y_{i} = c$.
Let $z$ denote the pre softmax probabilities of a neural network, and $f(x_{i,c};\mathbf{\theta})$ denote the pre-activation scores of the model output. The loss of the DNN is given by $\mathcal{L}( f(x_{i,c};\mathbf{\theta}),y_{i})$. We often use \Hess to denote the Hessian matrix $\mathcal{L}^{''} \in \mathbb{R}^{D \times D}$, where $D$ is the cardinality of $\theta$. We refer to the layerwise Hessian as \Hessl = $\frac{\partial^2 \mathcal{L}}{\partial \theta_{li} \partial \theta_{lj}}$ where $i,j \in \{1, ..., |\theta_l|\}$ and $|\theta_l|$ is the number of parameters in layer $l$. SGD stands for Stochastic Gradient Descent with a suitable minibatch size, and  $Tr(\mathbf{A})$ denotes the trace of matrix $\mathbf{A}$. 


\section{Layerwise Loss Landscape Analysis using the Hessian Eigenspectrum}
\label{sec_layerwise_loss}
DNN models largely follow a layerwise composition of functions and recent methods to improve performance (e.g. batch normalization) are also introduced at a layer level. It is hence natural to ask \textit{how the loss landscape at every layer behaves, especially when compared to the overall loss landscape}. 
We seek to address this question in this section.

The geometry of loss landscapes of DNN models is characterized by the Hessian matrix of the loss function, and our work in this section focuses on analyzing the Hessian eigenspectrum of each layer, in particular by analyzing their individual Gauss-Newton decompositions \cite{pmlr-v70-botev17a}. 
We begin with introducing the reader to the Gauss-Newton decomposition of the Hessian matrix, $\mathbf{Hess} = \mathbf{H} + \mathbf{G}$, defined in \cite{sagun2016eigenvalues,papyan2018full} where:
%
%
%
%
\begin{equation}
\label{eqn_H}
\mathbf{H} = \text{Ave}_{i,c} \Bigg{\{} \sum_{c'=1}^{C}\frac{\partial \mathcal{L}(z, y_c)}{\partial z_{c'}}\Bigg|_{z_{i,c}} \frac{\partial^2 f_{c'}(x_{i, c}; \theta)}{\partial \mathbf{\theta}^2} \Bigg{\}} 
\end{equation}
\begin{equation}
\label{eqn_G}
\mathbf{G} = \text{Ave}_{i, c}\Bigg{\{} \frac{\partial f(x_{i, c}; \theta)^T}{\partial \mathbf{\theta}} \frac{\partial^2 \mathcal{L} (z, y_c)}{\partial z^2}\Bigg|_{z_{i, c}} \frac{\partial f(x_{i, c}; \theta)}{\partial \mathbf{\theta}} \Bigg{\}}
\end{equation}
and $z_{i,c} = f(x_{i, c}; \theta)$, $\text{Ave}_{i, c}$ denotes the average across all observations $i$ and classes $c$. It is straightforward to note that the Gauss-Newton decomposition of the layerwise Hessian, $\mathbf{Hess}_l$, can be given by $\mathbf{G}_l + \mathbf{H}_l$ for $l = 1, \cdots, L$. This follows from the fact that the layerwise Hessian corresponds to blocks around the diagonal in the overall Hessian, and the corresponding Hessian is simply restricted to $\theta_l$, the parameters of that layer. Hence, the layerwise Hessian decomposition into \Gl and \Hl is given as: 
\begin{equation}
\label{eqn_G_layerwise}
\mathbf{G}_l = \text{Ave}_{i, c}\Bigg{\{} \frac{\partial f(x_{i, c}; \theta)^T}{\partial \mathbf{\theta}_l} \frac{\partial^2 \mathcal{L} (z, y_c)}{\partial z^2}\Bigg|_{z_{i, c}} \frac{\partial f(x_{i, c}; \theta)}{\partial \mathbf{\theta}_l} \Bigg{\}}
\end{equation}
\begin{equation}
	\label{eqn_H_layerwise}
	\mathbf{H}_l = \text{Ave}_{i,c} \Bigg{\{} \sum_{c'=1}^{C}\frac{\partial \mathcal{L}(z, y_c)}{\partial z_{c'}}\Bigg|_{z_{i,c}} \frac{\partial^2 f_{c'}(x_{i, c}; \theta)}{\partial \mathbf{\theta}_l^2} \Bigg{\}}    
\end{equation}
We study the eigenspectrum of the layerwise Hessian, \Hessl, through the eigenspectra of each \Gl and  \Hl. Among the tools available to efficiently compute the eigenspectrum, we use the Lanczos method \cite{Lanczos:1950zz} owing to its better performance in time incurred over competing methods such as KPM \cite{lin2016approximating} in earlier efforts such as \cite{papyan2018full}. The Lanczos method computes the eigenspectrum of a symmetric matrix by reducing it to a tridiagonal form, $\mathbf{T}_D \in \mathbb{R}^{D \times D}$, and computing the spectrum of $\mathbf{T}_D$ instead. The original Lanczos method however requires an inner iterative loop (for $D$ iterations) to reorthogonalize the obtained vectors in each iteration due to numerical errors. For matrices with very large $D$ such as in DNNs, this orthogonalization step is computationally intensive and necessitates the simultaneous use of multiple high-end GPUs (such as in \cite{eigval2019ghorbani}) for implementation, making it impractical. 

To overcome this issue, \cite{lin2016approximating} proposed an approximation where the Lanczos method (Algorithm details in supplementary section) is used only for $M << D$ iterations, thus obtaining $M$ eigenvalue-eigenvector pairs, and subsequently estimating the eigenspectrum density using a Gaussian convolution on the $M$ outputs of the approximate Lanczos. In particular, this method is based on writing the eigenspectrum of any large matrix as $\phi(t) = \frac{1}{D} \sum_{i=1}^{D} \delta(t-\lambda_i)$ where $\delta$ is the Dirac delta function, $\lambda_i$ is the $i^{th}$ eigenvalue, $D$ is the total number of eigenvalues, and $\phi(t)$ is the frequency of eigenvalue $t$. Instead of computing the entire spectrum, this approximation computes a Gaussian density convolution $\phi_\sigma (t) = \frac{1}{D} \sum_{i=1}^{D} \mathbf{y}_i[1]^2 g_{\sigma}(t-\lambda_i)$, where $g_{\sigma}(t-\lambda_i)$ is a Gaussian centered at $\lambda_i$ with width $\sigma$. For more details of this method, the interested reader is requested to refer to \cite{lin2016approximating,papyan2018full}.

Earlier work that analyzed the entire loss landscape of DNN models using the Hessian eigenspectrum have made interesting observations. In particular, \cite{sagun2017empirical} as well as \cite{papyan2018full} observed that the Hessian eigenspectrum is divided into a \textit{bulk} region, and an \textit{outlier} region. More interestingly, both \cite{sagun2017empirical} and \cite{papyan2018full} also showed that the number of outliers in the Hessian eigenspectrum is approximately the number of classes, $C$. With this background in context, we study the layerwise loss surface using the Hessian eigenspectrum, as computed using Lanczos method. We conducted studies on many state-of-the-art DNN models including VGG11,13,16 (with and without Batch Normalization), ResNet18 and DenseNet on MNIST, FashionMNIST, and CIFAR10 datasets. Due to space constraints, we report our results only with VGG11-BN on CIFAR-10 in Figure \ref{fig_spectrum_vgg11nb_cifar10} and the remaining results can be found in the Supplementary material. We inferred similar observations for all of these models however, and all our results can be reproduced using our anonymized source code repository shared herewith\footnote{\url{https://anonymous.4open.science/r/6eaea11b-879c-43a8-992b-ddb3c95b23c8/}}.
\begin{figure}
\centering
    \includegraphics[width=\linewidth]{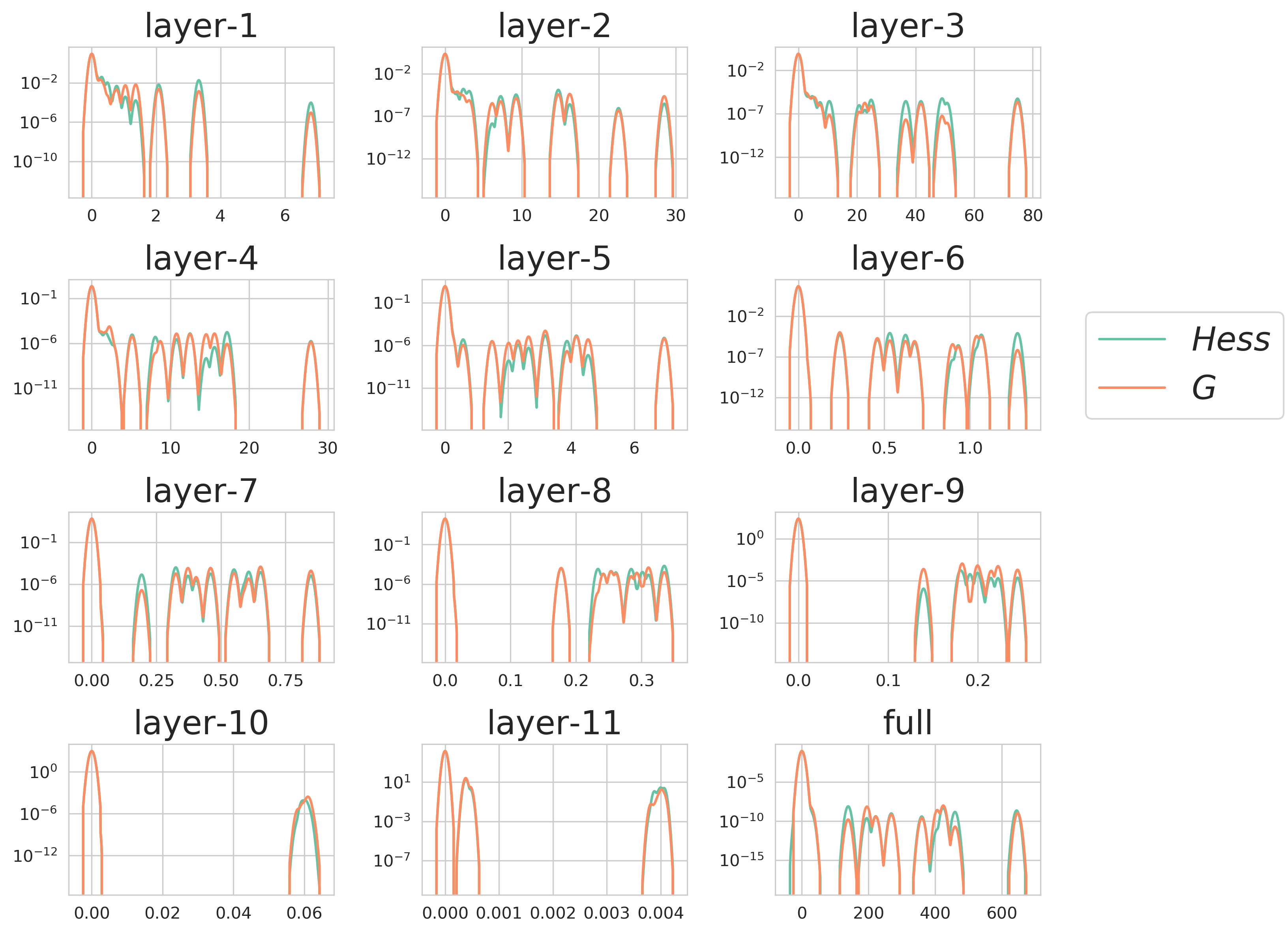}
    \caption{Eigenspectra of \HessL and \GL on VGG11+BatchNormalization model trained on CIFAR-10 (Best viewed in color). Note that where the graph of the Hessian is not visible, the spectra of Hessian and G overlap completely. The last subplot ``full'' refers to eigenspectra of the \Hess and \G of the entire network;}
    \vspace{-0.5cm}
    \label{fig_spectrum_vgg11nb_cifar10}
\end{figure}

Figures \ref{fig_spectrum_vgg11nb_cifar10}  show the eigenspectrum of \HessL and \GL of VGG11(with Batch Normalization), ResNet18 when trained on the CIFAR10 dataset. The eigenspectrum of \GL is obtained by the decomposition of \HessL defined in Equations \ref{eqn_G_layerwise} and \ref{eqn_H_layerwise}. We do not report the spectrum of \Hl, since we found the eigenvalues to be mostly $\approx$ 0. This was pointed out by \cite{sagun2017empirical} in their work too, who observed that the spectra of $H \approx 0$ and that of $Hess \approx G$. This is reflected in our results; for e.g., in Figures \ref{fig_spectrum_vgg11nb_cifar10}, one can see that the spectra of \HessL and \GL almost overlap across all layers. Based on our results across models and datasets including Figure \ref{fig_spectrum_vgg11nb_cifar10}, we report our inferences below.

\cite{papyan2018full} (cf. Fig 6) reported that the spectra of \G peaks at a particular epoch during training, where the number of outliers in the eigenspectra of $\mathbf{G}$ and $\mathbf{Hess}$ is $C$, the number of classes. Interestingly, we too find that the number of outliers in the layerwise eigenspectra of \GL and \HessL are also $C$, across almost all layers. One can notice this on careful observation in Figure \ref{fig_spectrum_vgg11nb_cifar10} (notice the number of peaks in the spectra in each subplot). 



\begin{figure}[h]
    \centering
    \includegraphics[width=0.9\linewidth]{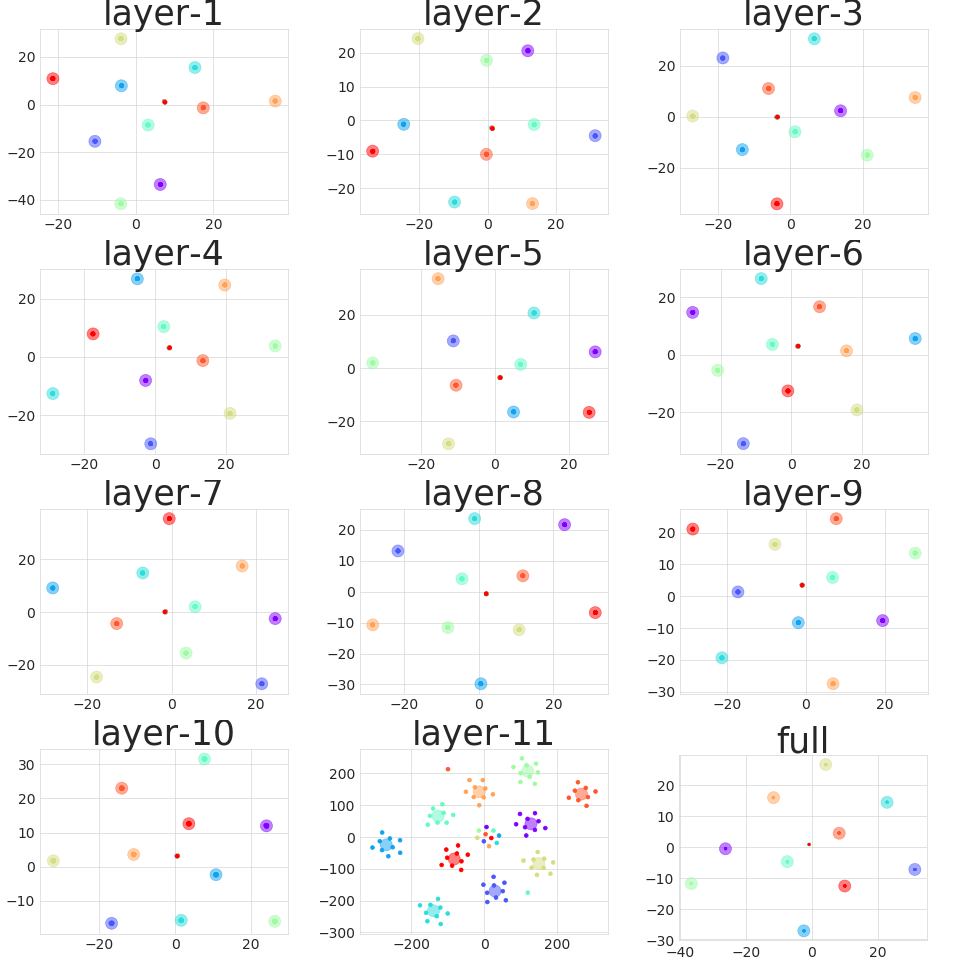}
    \caption{Layerwise two-dimensional t-SNE embeddings of $\delta_{c}$, obtained as a decomposition of \GL in \protect \cite{papyan2018full} show $C$ clusters in each subplot (VGG11+BN on CIFAR10). The last subplot ``full'' corresponds to $\delta_{c}$ of the entire \G matrix.}
    \vspace{-0.5cm}
     \label{fig_tsne_layerwiseG}
\end{figure}

More recently, \cite{papyan2019measurements} showed that $\mathbf{G}$ can be written as $\mathbf{G} = \frac{1}{n} \Delta \Delta^T = \text{Ave}_{i,c} \Big\{ \sum_{c'=1}^C \delta_{i,c,c'}\delta_{i,c,c'}^T\Big\}$, where $\delta_{i,c,c'}$ is the $c'$-th column of a submatrix of $\Delta$ (please see Sec 2, Eqns 15-16 of \cite{papyan2019measurements} for details). It was further shown that the t-SNE plots of $\delta_{c}$ (obtained by averaging $\delta_{i,c,c'}$ over all observations $i$ and all potential classes $c'$) yields $C$ clusters again. We conducted these studies layerwise, and found that the same observation holds for the layerwise \GL too, as shown in Figure \ref{fig_tsne_layerwiseG}, where we plot the t-SNE embeddings of $\delta_c$ obtained by decomposing \GL matrix for all layers in a DNN.

\begin{figure}
  \begin{center}
    \includegraphics[scale=0.37]{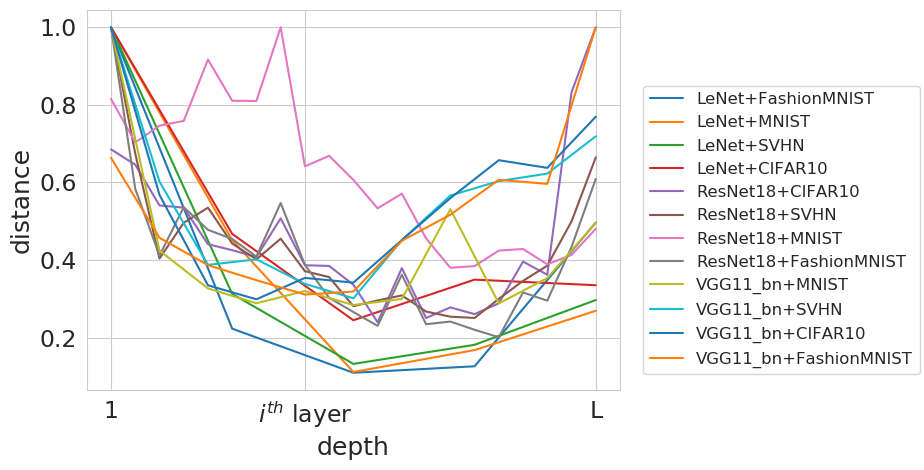}
    \caption{Plots of normalized Wasserstein distance between the spectra of $\mathbf{Hess}_l; \forall l = 1 \cdots L$ and $\mathbf{Hess}$ of different DNN models across datasets.}
    \vspace{-0.5cm}
     \label{fig:match-distros}
  \end{center}
\end{figure}

To understand further, we studied if there are particular layers in these well-known, often-used DNN models whose loss landscape matches the overall loss surface the most. We considered the eigenspectrum density of layers, and the entire network, and computed the normalized Wasserstein distance \cite{wasser}:
between the spectra of \HessL and \Hess for different DNN models. These results are shown in Figure \ref{fig:match-distros}. It can be clearly observed that the spectra of \Hessl of middle layers of DNN models are very similar to spectrum of the overall \Hess. Such an observation has not been made hitherto, to the best of our knowledge. We also conducted this study using other measures such as KL-divergence (results in Supplementary section) which had a similar observation about the middle layers being closest to the overall loss surface.
\vspace{-0.5cm}
    

\noindent \paragraph{Summary of Observations:} Based on our results, we proposition that the behavior of outliers in the layerwise Hessian eigenspectra, especially the grouping into $C$ clusters, indicates that every layer of state-of-the-art DNN models encapsulates discriminative capability, i.e., the capability to discriminate between the classes. Further, the loss landscape of the middlemost layers of DNN models consistently match the overall loss surface. These new observations on the layerwise understanding on the loss surfaces, especially the influence of middle layers on the loss surface, can potentially be used in regularization methods, use of batch-normalization, or in other training methods.
\section{Evolution of Hessian Eigenspectra}
\label{sec_evolution_spectra}
Continuing the discussion from the previous section, we now shift our focus towards analyzing the eigenspectra over the course of DNN training. 
There have been a few efforts in the recent past by other researchers \cite{chaudhari2018stochastic,eigval2019ghorbani} seeking to understand gradient descent trajectories. \cite{chaudhari2018stochastic} showed the relationship between SGD and variational inference while studying these trajectories; 
In a related attempt, \cite{eigval2019ghorbani} studied GD trajectories by studying the Hessian eigenspectra, but their focus was largely on the tool developed by them to analyze the Hessian than present any newer observations. The work closest to ours in this direction is that of \cite{sharpest} where the authors study the trajectories of SGD and the connection to generalization performance. In particular, they show the behavior of SGD along sharp directions of the loss surface, and conclude that a variant of SGD can find sharp minima with good generalization. We are fundamentally different from this work from multiple perspectives: (i) Their work follows earlier efforts in \cite{dinh}, which show that sharp minima can be generalizable too, while our work follows most earlier efforts such as \cite{keskar2017} which show the importance of flatter minima in generalization; (ii) Their work relies on experiments on a simple 4-layered CNN on CIFAR-10, while we use state-of-the-art DNN models; (iii) We also introduce a tool to efficiently compute the trace of the Hessian and analyze the spectra, which has not been used before. While the abovementioned literature in this field have different perspectives and our work was conceived independent of these efforts, the presence of these efforts only support the need for such analysis in the community.

We begin by studying the evolution of the eigenspectra for the entire Hessian, and subsequently study the layerwise Hessians. To this end, we use the maximum eigenvalue, $\lambda_{max}$, and the trace of the Hessian, considering it is not trivial to study the entire spectra over all epochs of training. We note that both these quantities, $\lambda_{max}$ and trace of Hessian denoted by $Tr(\mathbf{\mathcal{L}^{''}})$, provide understanding of the curvature of the loss surface (higher these values, steeper the curvature). Clearly, the reduction in both these quantities indicates flatter regions of the loss surface. While $\lambda_{max}$ can be obtained using the methods discussed in Sec \ref{sec_layerwise_loss}, computing the trace of the Hessian of very large matrices (as in DNNs) is computationally non-trivial. Earlier work that attempted such a direction \cite{sharpest} did not take into account the complete spectrum for this reason. We hence introduce the use of tools from randomized numerical linear algebra, in particular the Hutchinson method, to compute the trace of the Hessian without explicit computation of the Hessian \cite{avron2011randomized}. The trace is obtained as:  
 \begin{align*}
 Tr(\mathbf{\mathcal{L}^{''}}) =Tr(\mathbf{\mathcal{L}^{''} I}) =  Tr(\mathbf{{\mathcal{L}^{''}} \mathbb{E}[\mathbf{v} \mathbf{v}^\top]}) \\ =  \mathbb{E}[Tr( \mathbf{\mathcal{L}^{''}} \mathbf{v} \mathbf{v}^\top)] =\mathbb{E}[ \mathbf{v}^\top\mathbf{\mathcal{L}^{''}}\mathbf{v}]
 \end{align*} 
where $\mathbf{v} \sim \mathcal{N}(0, \mathbf{I})$.
\begin{algorithm}[t]
\caption{Hutchinson method to compute Hessian trace}
\label{algo_hutchinson}
\begin{flushleft}
\textbf{Input}: Model parameters: $\theta$, Loss function $\mathcal{L}$; No of iters: n \\
\textbf{Output}: Trace of $\frac{\partial ^ 2 \mathcal{L}}{\partial \theta^2}$
\end{flushleft}
\begin{algorithmic}[1] 
\STATE trace = 0
\FOR{t = 1 $\cdots$ n}
\STATE $\mathbf{v} \sim \mathcal{N}(0,\mathcal{I})$; $ g_\theta = \frac{\partial \mathcal{L}}{\partial \theta}$;  trace = trace + $\mathbf{v}^\top \frac{\partial (g_\theta^T \mathbf{v})}{\partial \theta}$; 
\ENDFOR
\STATE \textbf{return} trace/n
\end{algorithmic}
\vspace{-2pt}
\end{algorithm}
The equality $  \mathbb{E} [\mathbf{v} \mathbf{v}^\top] = \mathbf{I}$ comes from the expectation of quadratic form when $\mathbf{v} \sim \mathcal{N}(0,\mathbf{I})$. We can also draw $\textbf{v}$ from the Rademacher distribution where each entry is either $+1$ or $-1$ with  probability 0.5. 
\begin{figure}
  \begin{center}
    \includegraphics[scale=0.35]{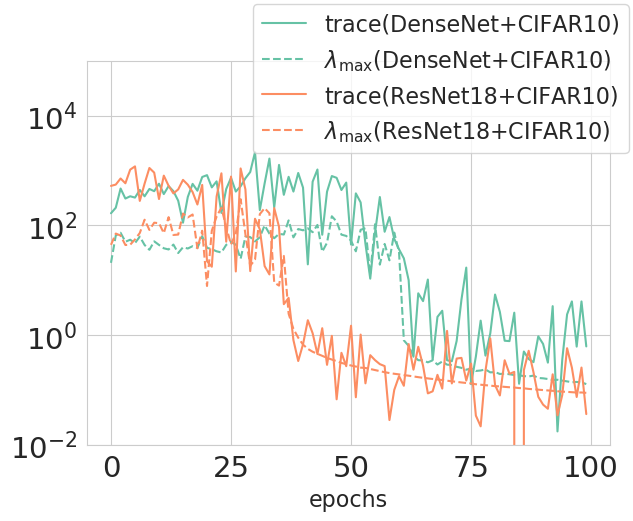}
    \vspace{-8pt}
    \caption{Evolution of $\lambda_{\max}$, trace of Hessian}
     \vspace{-12pt}
    \label{fig_trace_evolution_cifar10}
  \end{center}
\end{figure}
The complete methodology to compute $Tr(\mathbf{\mathcal{L}^{''}})$ using the Hutchinson method is summarized in Algorithm \ref{algo_hutchinson}.

Figure \ref{fig_trace_evolution_cifar10} shows the plot of $\lambda_{\max}$ and trace of Hessian for Resnet-18 and DenseNets models on the CIFAR-10 dataset. 
A clear observation, which was also reflected in all our experiment runs, is that both $\lambda_{\max}$ and the $Tr(\mathbf{\mathcal{L}^{''}})$ reduce over epochs, pointing to the inference that the curvature becomes smaller over training, thus leading to flatter minima. 
\begin{figure*}[ht]
  \subfloat[VGG11 and VGG11+BN (MNIST)]{
	\begin{minipage}{
	   0.33\textwidth}
	   \centering
    \includegraphics[scale=0.3]{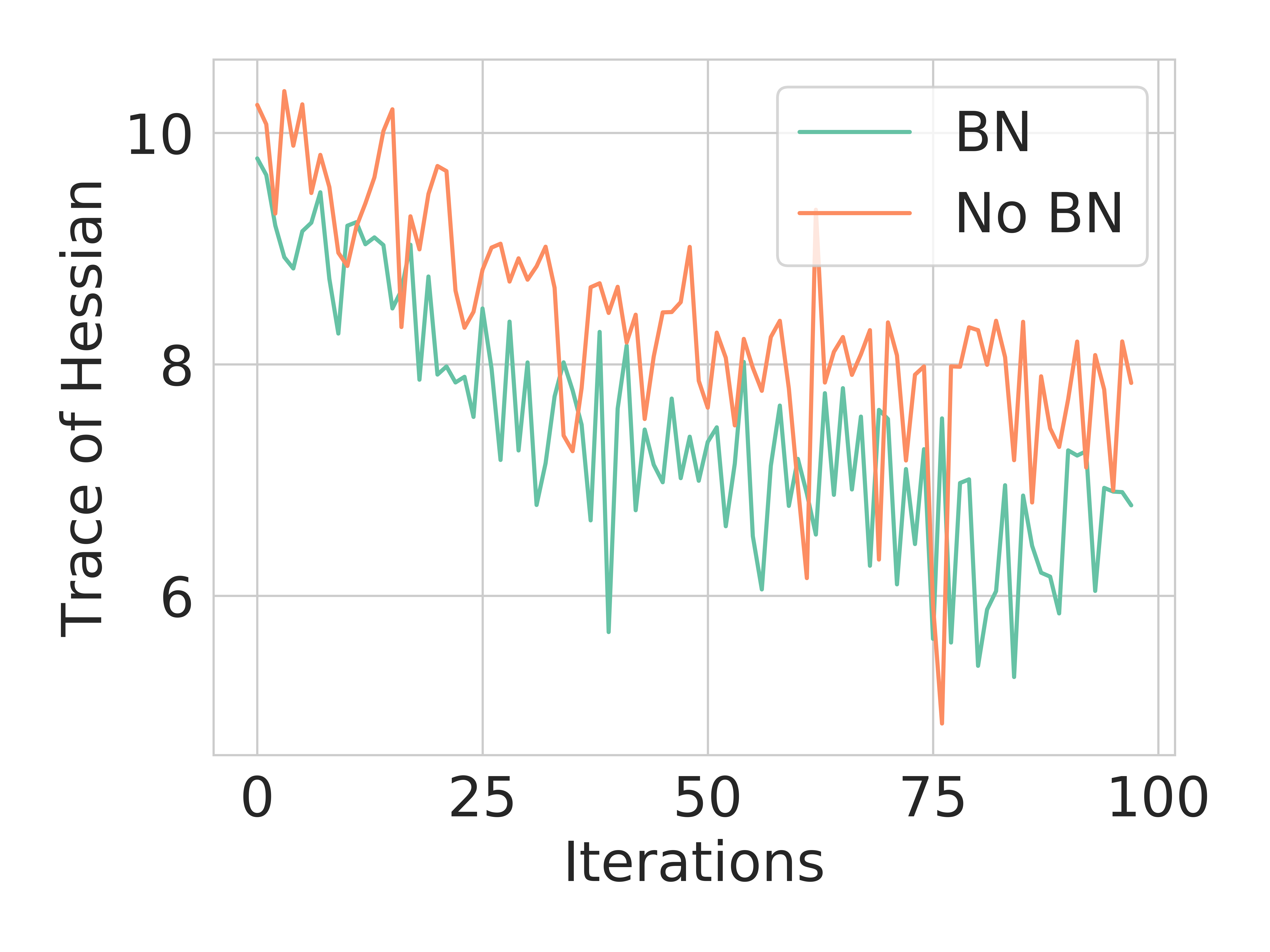}
	\end{minipage}}
\subfloat[ResNet18 (skip,no-skip weights) (MNIST)]{
	\begin{minipage}{
	   0.33\textwidth}
	   \centering
    \includegraphics[scale=0.3]{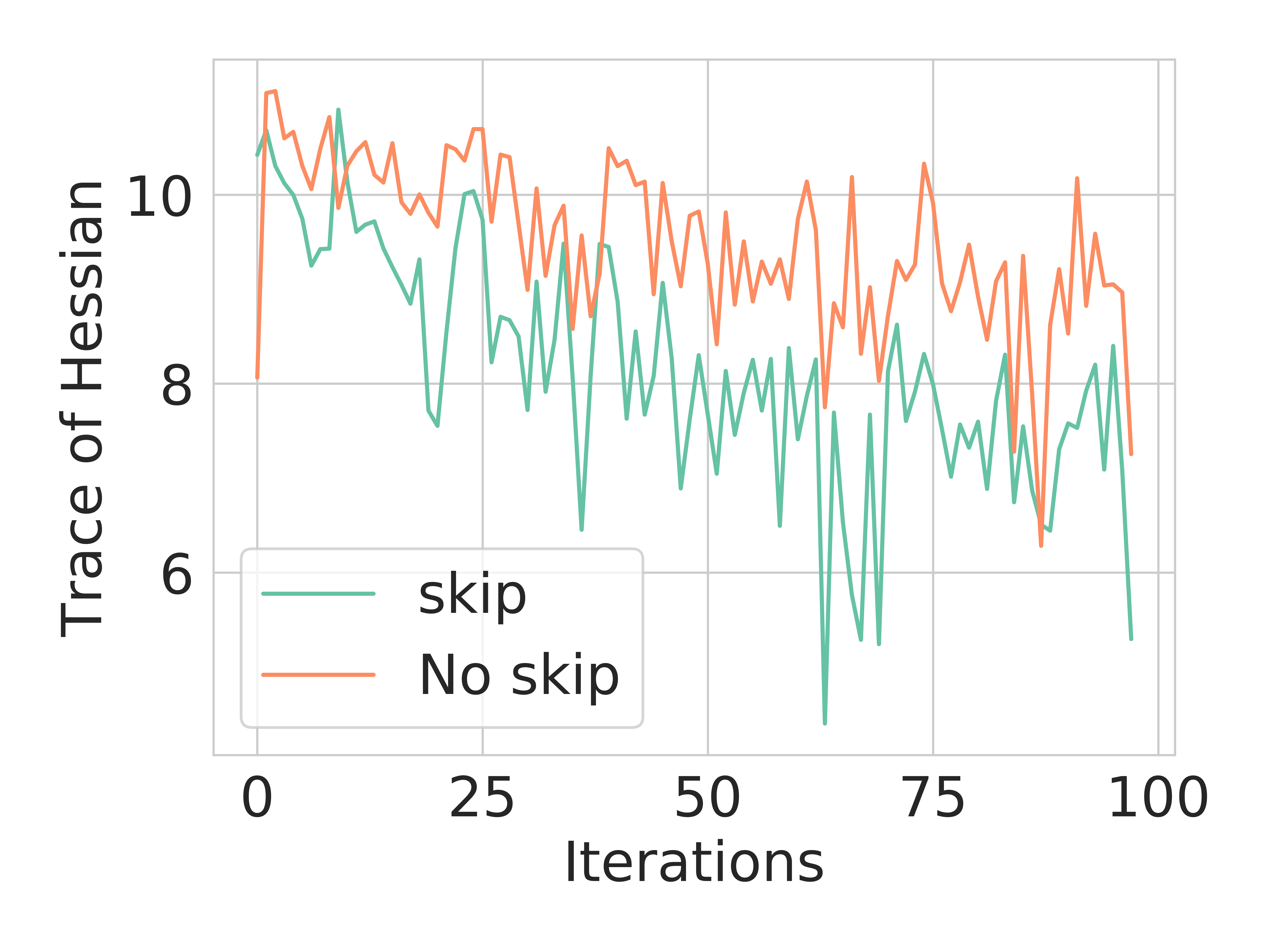}
	\end{minipage}}
  \subfloat[VGG13 and VGG13+BN (CIFAR10)]{
	\begin{minipage}{0.33\textwidth}
	   \centering
    \includegraphics[scale=0.3]{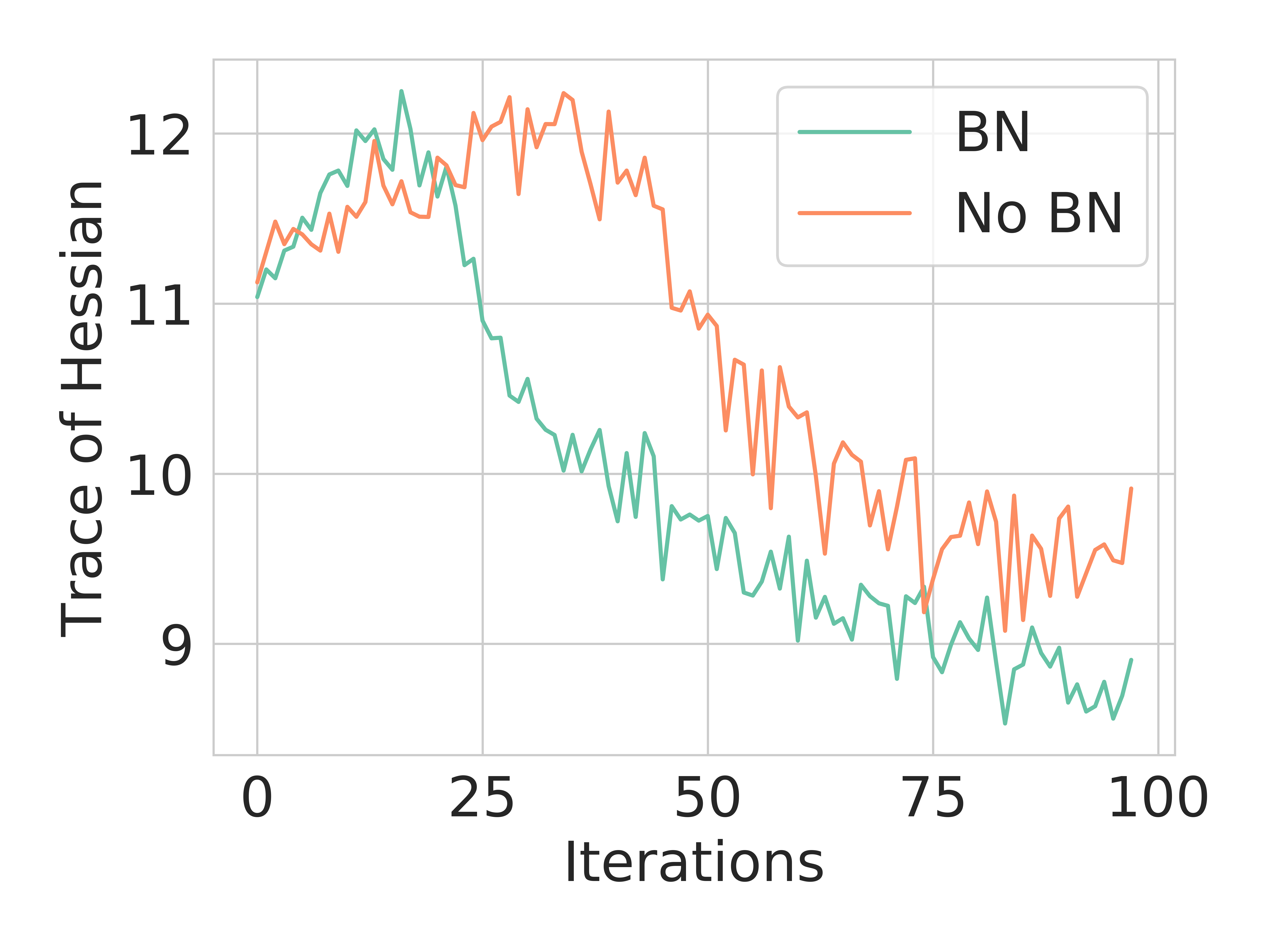}
	\end{minipage}}
\caption{Evolution of trace of Hessian of state-of-the-art models with and without batch normalization/skip connections on MNIST and CIFAR10. Training accuracy was close to state-of-the-art for all considered models at the end of these iterations.}
\vspace{-6pt}
\label{fig_trace_batchnorm}
\end{figure*}

To further understand the connection between the trace of the Hessian and generalization performance, we explicitly considered architectural changes in models that have improved generalization performance in recent years. In particular, we used batch normalization and skip connections (such as in ResNets and DenseNets), which have proven improvements in generalization performance over the last few years. Figure \ref{fig_trace_batchnorm} shows the results of these experiments.
In addition to corroborating our above inference that the trace of the Hessian decreases over training, these plots also show that the use of batch normalization plays an important role in reducing the curvature, as well as the inference that the Hessian trace and generalization performance are closely linked, as evident from these figures. Similar trends on DenseNets and are provided in supplementary material.
\begin{figure}
\begin{center}
    \includegraphics[scale=0.4]{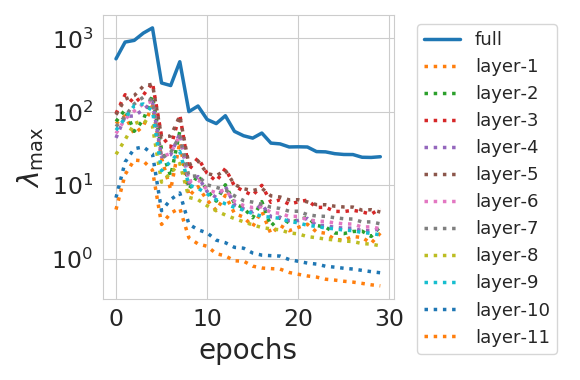}
    \caption{Evolution of $\lambda_{max}$ of layerwise Hessian on VGG11+MNIST}
    \label{fig_lmax_layerwise}
  \vspace{-0.5cm}
  \end{center}
\end{figure}

We subsequently studied the evolution of the layerwise Hessian eigenspectra in our experiments, and report one result (owing to space constraints) in Figure \ref{fig_lmax_layerwise}. Clearly, the same trends hold for $\lambda_{max}$ of each layerwise Hessian, where the value goes down over training. We could not include the trace results to avoid overcrowding in the graph, but trace values of layerwise Hessians showed a similar trend. Interestingly, we once again see that middle layers have values of $\lambda_{max}$ closest to that of the entire network across the epochs. This was a consistent observation across our studies, and points to a deeper connection which is yet to be theoretically understood. We hope that this work will raise this pertinent question in the community.
\vspace{-0.3cm}
    

\noindent \paragraph{Summary of Observations:} Our study of the evolution of Hessian eigenspectra over training presents a few important observations: (i) Our results broadly support earlier work, such as \cite{keskar2017}, that state-of-the-art DNN models converge to flatter minima (with lower curvatures on the loss surface); (ii) there is an evident connection between the trace of the Hessian and generalization error, as shown by our studies with and without batch normalization and skip connections; and (iii) the same trend holds for the layerwise Hessian too, and importantly, the middlemost layers have $\lambda_{max}$ closest to that of the full network across the training. 

Motivated by these observations, we now present a new regularization method, layerwise Hessian Trace Regularization, which seeks to lower the layerwise Hessian trace during training explicitly, with an aim to improve generalization performance. 
\section{Layerwise Hessian Trace Regularization}
\label{sec_hessian_trace_regularization}
The proposed layerwise regularization method for DNNs, which we call layerwise Hessian Trace Regularization (or HTR), is motivated by our empirical observations that state-of-the-art models reduce the trace of layerwise Hessian over training. Importantly, we show that this layerwise regularization approach lends itself to an interesting premise - that regularizing only on the trace of Hessian of the middlemost layers by itself provides strong performance. This observation is in alignment with findings from our earlier sections. To the best of our knowledge, such a layer-specific regularization method has not been studied before.

We modify the DNN training objective as:
\begin{equation}
\label{eqn_htr_objective}
    \mathbf{\theta}^* = \arg \min_{\theta} \mathcal{L} ( f(x;\theta) ), y) + \gamma \sum_{l=1}^{L}  \text{Tr}(\mathbf{Hess}_l )
\end{equation}
\noindent where $ \text{Tr}(\mathbf{Hess}_l)$ is the trace of the Hessian of the loss function at layer $l$ and $\gamma$ is a regularization hyperparameter. The proposed layerwise regularizer works by penalizing the sum of the trace of layerwise Hessians. We choose a uniform weighting of layers in this work. It is possible to weight Hessians of different layers differently, which is an interesting future direction of work. This penalization at every layer helps encourage SGD to converge to solutions with flatter minima, and hence generalize better \cite{keskar2017}. We studied the usefulness of this regularizer on several well-known DNN models across multiple datasets (MNIST, FMNIST, SVHN, CIFAR10, CIFAR100). Cross-entropy loss, $\mathcal{L}_{ce}$, was used to train each of these classification models.
We used a momentum of 0.9, learning rate of 1e-2, L2 regularization co-efficient of 1e-3, batch size of 64. We trained our model on a NVidia GeForce GTX 1080 GPU with 12GB GPU memory. We ran 5 trials of each experiment to avoid bias in randomness of initialization. Our anonymized source code (Python) is available \footnote{\url{https://anonymous.4open.science/r/6eaea11b-879c-43a8-992b-ddb3c95b23c8/}} for reproducibility.

\begin{table}[]
\begin{tabular}{|l|l|l|}
\hline \hline
\multicolumn{1}{|c|}{\textbf{Model+Dataset}} & \multicolumn{1}{c|}{\textbf{\begin{tabular}[c]{@{}c@{}}$\mathcal{L}_{ce}$+HTR\\ \end{tabular}}} & \multicolumn{1}{c|}{\textbf{$\mathcal{L}_{ce}$}} \\ \hline \hline

LeNet + MNIST       & \textbf{ 1.19 $\pm$ 0.07} &  1.25 $\pm$ 0.03    \\ \hline
LeNet + FMNIST      & \textbf{11.14 $\pm$ 0.04} & 11.22 $\pm$ 0.11    \\ \hline
LeNet + SVHN        & \textbf{10.77 $\pm$ 0.26} & 10.94 $\pm$ 0.06    \\ \hline
VGG11 + CIFAR10     & \textbf{15.11 $\pm$ 0.22} & 18.20 $\pm$ 0.45    \\ \hline
VGG13 + SVHN        & \textbf{7.47 $\pm$ 0.01}  & 7.73 $\pm$ 0.28     \\ \hline
ResNet18 + CIFAR10  & \textbf{11.95 $\pm$ 0.22} & 11.97 $\pm$ 0.14    \\ \hline
VGG11-BN + CIFAR100 & \textbf{44.98 $\pm$ 0.05} & 45.25 $\pm$ 0.44    \\ \hline
\hline
\end{tabular}
\caption{Comparison of generalization error when trained using cross-entropy loss ($\mathcal{L}_{ce}$) with and without HTR}
\vspace{-5pt}
\label{tab_hrt_momentum_results}
\end{table}

\begin{table}[]
\begin{tabular}{|l|l|l|}
\hline \hline
\multicolumn{1}{|c|}{\textbf{Model+Dataset}} & \multicolumn{1}{c|}{\textbf{Middle Layers}} & \multicolumn{1}{c|}{\textbf{Full Network}} \\ \hline \hline
LeNet + MNIST                    & \textbf{ 1.18 $\pm$ 0.07}   &  1.19 $\pm$ 0.07    \\ \hline
LeNet + FMNIST                   & \textbf{11.10 $\pm$ 0.07}   & 11.14 $\pm$ 0.04    \\ \hline
ResNet18 + CIFAR10               & \textbf{11.87 $\pm$ 0.12}   & 11.95 $\pm$ 0.22    \\ \hline
VGG11-BN + CIFAR100              & \textbf{44.81 $\pm$ 0.15}   & 44.98 $\pm$ 0.05    \\ \hline 
\hline
\end{tabular}
\caption{Comparison of generalization error when HTR is used only on middle layers, versus HTR on all layers}
\vspace{-0.5cm}
\label{tab_middle_htr}
\end{table}

\begin{table*}[h]
	\begin{tabular}{|l|l|l|l|l|l|}
		\hline \hline
		\multicolumn{1}{|c|}{\textbf{LeNet+MNIST}} & \multicolumn{1}{c|}{\textbf{LeNet+FMNIST}} & \multicolumn{1}{c|}{\textbf{LeNet+SVHN}} & \multicolumn{1}{c|}{\textbf{VGG11+CIFAR10}} & \multicolumn{1}{c|}{\textbf{ResNet18+CIFAR10}} & \multicolumn{1}{c|}{\textbf{VGG11-BN+CIFAR100}} \\ \hline \hline
		1.15 $\pm$ 0.02   &11.11 $\pm$ 0.05 &  9.96 $\pm$ 0.10          &15.08 $\pm$ 0.16        & 11.89 $\pm$ 0.13          &   44.32 $\pm$ 0.29              \\ \hline \hline
	\end{tabular}
	\caption{Generalization error, with HTR objective is incorporated into L2 regularizer. (Baseline comparision in Table \ref{tab_hrt_momentum_results}) }
	\vspace{-0.6cm}
	\label{tab-htr-l2}
\end{table*}

\begin{figure}[h]
    \centering
    \includegraphics[scale = 0.53]{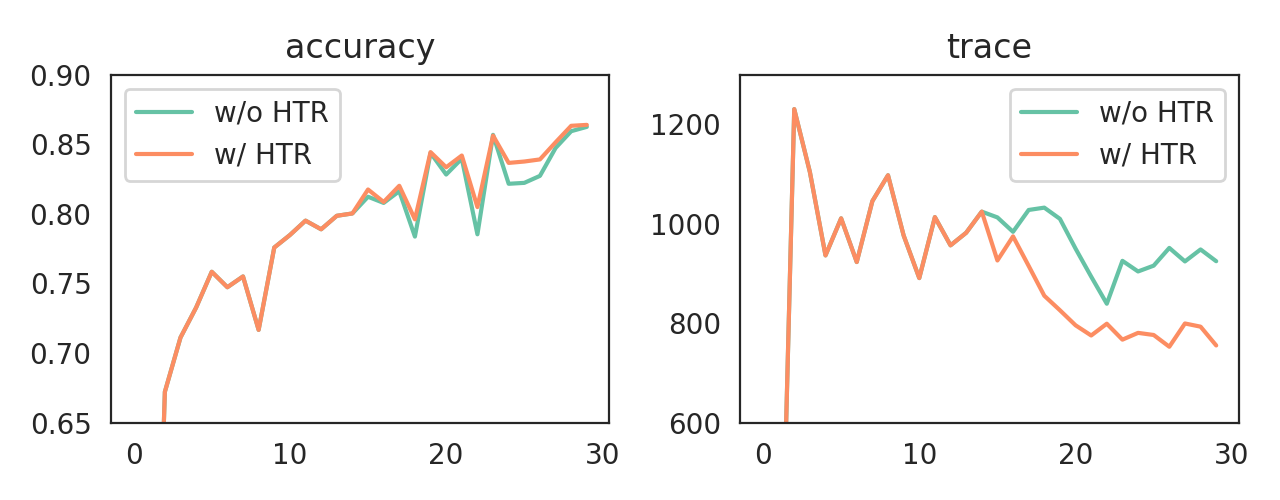}
    \caption{Plots of training accuracy and Hessian trace with and without HTR (LeNet+FMNIST); $x$-axis corresponds to training epochs.}
    \vspace{-5pt}
    \label{fig_htr_accuracy_trace_plot}
\end{figure}

Table \ref{tab_hrt_momentum_results} reports the generalization error when proposed HTR is incorporated into the cross-entropy loss objective. It can be clearly noticed that generalization error is better when HTR is incorporated into the objective of the DNN training across the considered datasets and models. All our experiments were conducted in a fair manner by tuning regularization hyperparameters in each of these methods and reporting the best result for each of the considered methods. 
This improvement in generalization performance happens with almost no change in training accuracy, thus showing promise in reducing the generalization gap. Fig \ref{fig_htr_accuracy_trace_plot} shows the training accuracy with and without HTR. This figure also shows the sum of traces of layerwise Hessians, 
which is indeed lower at the end of training with the proposed HTR method, suggesting solutions corresponding to relatively flatter minima. 


Building further on the observations from the earlier section, where the statistical properties of the loss surface of the middle layers were found to be closest to the loss surface of the overall network, we leveraged the layerwise nature of the proposed HTR to penelize the middlemost layers alone. Such an approach offers computational advantages too, since the number of parameters involved in that term is restricted now to weights only from the middlemost layers. 
Table \ref{tab_middle_htr} reports the generalization error results on different datasets when trained only by penalizing the middlemost layers, against penalizing all layers. We considered the layers between $\frac{1}{4}^{th}$ and $\frac{3}{4}^{th}$ of the total number of layers as ``middle layers'' for purposes of these experiments. It is interesting to note that the generalization error tht the models in fact perform marginally better in this case, suggesting a deeper connection between the middlemost layer and the overall network's error surfaces. 

Continuing further, we note that the proposed HTR is a general idea, and can also be used in conjunction with other regularizers such as L2 weight decay. We experimented further to study how the proposed HTR works when combined with standard L2 regularizer and the results are reported in Table \ref{tab-htr-l2}. It can be noted that the generalization error has reduced further in comparison to using HTR alone (see Table \ref{tab_hrt_momentum_results} for the baseline), indicating the effectiveness of HTR when used with other regularizers.

\begin{figure}[h]
   \begin{center}
     \includegraphics[width=0.85\linewidth]{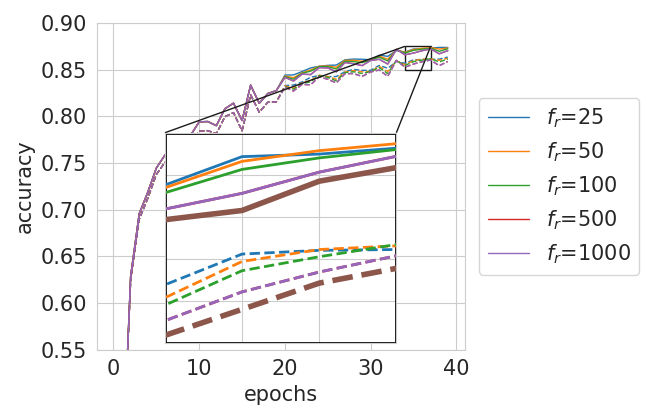}
     \caption{Ablation study on update frequency $f_r$ on LeNet + FMNIST (Best viewed in color, solid lines = train accuracy, dashed lines = test accuracy)}
   \label{fig_htr_freq_study}
   \end{center}
 \end{figure}

One issue with the proposed regularizer is that computing the Hessian trace or its derivative at every step can be computationally intensive, although this is mitigated to an extent using layerwise Hessians (whose sizes are smaller). However, to address this issue, we propose the use of the HTR term only at periodic intervals over the training process, and not at every iteration. We term this the penalization/update frequency, $f_r$. We studied further impact of the choice of $f_r$, and report these results in Figure \ref{fig_htr_freq_study}. As evident from the figure, even lower values of the update frequency performed quite well, thus reducing the computational cost of this method. (The results in Table \ref{tab_hrt_momentum_results} were, in fact, obtained at $f_r = 50$).

Analyzing further, it may be a natural question to ask about understanding the difference between regularizing over the trace of the entire Hessian versus the sum of the trace of layerwise Hessians in the proposed method. We note that under the conditions studied in this work for Eqn \ref{eqn_htr_objective}, where we do not weight each layer differently, both are equivalent since the sum of traces of layerwise Hessians is the trace of the overall Hessian.  This is the reason for our result in Figure \ref{fig_htr_accuracy_trace_plot}, where the trace of the overall Hessian reduces over training with our regularizer. This can also be seen in Figure \ref{fig_lmax_layerwise}, where the trace of layerwise Hessian reduces, and the overall trace also reduces. There is, however, a considerable computational advantage with layerwise Hessian trace penalization when compared to full Hessian trace penalization. The trace of each layerwise Hessian can be computed in parallel leading to an increase in time efficiency by a factor of $L$ (number of layers in the neural network). Layerwise penalization is also effective from a memory perspective, since it is a smaller matrix to compute and store. A major reason for the lack of progress in using second-order information in training DNNs is: (i) The Hessian is computationally intensive to obtain, especially for large models with millions of parameters; (ii) Storing the Hessian matrix is also not memory-efficient. The proposed layerwise HTR, which is the first layerwise regularizer to the best of our knowledge, mitigates both these problems. 
There is now a considerably smaller memory footprint as we consider only a submatrix of the overall Hessian. The method also allows different co-efficients for penalizing different layers differently, or even ignoring certain layers, which could be an added advantage and we leave for future work. 

\section{Conclusion}
\label{sec_discussions_conclusion}

The layerwise analysis of loss surfaces of DNN models deserves the attention of the deep learning community. In this work, we analyzed how layerwise landscape loss properties correlate with overall loss landscape by studying properties primarily through the lens of the Hessian eigenspectrum. We analyzed both the spectral density, as well as specific properties such as maximum eigenvalue and trace of the Hessian over the training of these DNN models. Our studies on state-of-the-art models across datasets show that each layer of a DNN also maintains class discriminability, with the middlemost layers having the strongest connection to the overall loss surface. Our study of the evolution of the spectra showed that state-of-the-art DNN models seek flatter minima, and that middlemost layers maintain a relationship with the overall network through the training process. Motivated by this observation, we propose a new layerwise Hessian-based regularizer, Hessian Trace Regularization method that works promisingly when models and datasets become complex. We believe that the observations presented in this work will help deepen the community's understanding about DNN models in general.

\section{Acknowledgements}
This work has been partly supported by the funding received from DST, Govt of India, through the MATRICS program (MTR/2017/001047), MHRD and the Intel India PhD Fellowship. We also acknowledge IIT-Hyderabad and JICA for provision of GPU servers for the work. We thank the anonymous reviewers for their valuable feedback that improved the presentation of this work.
\bibliography{sample-base}


\newpage

\section{Supplementary Material}

In this section, we include more details on how to estimate the eigenspectrum of Hessian matrix using the Lanczos method, and more experimental results on the eigenspectra of state-of-the-art networks such as ResNet and DenseNet, which could not be included in the main paper due to space constraints.

\subsection{Eigenspectrum Estimation Algorithm:} 
Algorithm \ref{algo_approx_hessian_espectrum} presents the Lanczos method which estimates the eigenspectrum of a given matrix. It should be noted that the Lanczos method requires only Hessian-vector product (lines 5,7 in Algorithm) which can be easily approximated for a deep neural network. 

\begin{algorithm}[h]
\caption{Eigenspectrum Estimation of Hessian using Lanczos method \protect \cite{Lanczos:1950zz,lin2016approximating,papyan2018full}}
\label{algo_approx_hessian_espectrum}
\begin{flushleft}
\textbf{Input}: Hessian $\mathbf{A}$ with spectrum in range [-1, 1], Num of iterations $M$, Num of samples $K$, $\kappa$\\
\textbf{Output}:(Approx) Eigenspectrum density of $\mathbf{A}$
\end{flushleft}
\begin{algorithmic}[1] 
\FOR{i = 1, $\cdots$, M}
    \IF{i == 1}
        \STATE sample $\mathbf{v} \thicksim \mathcal{N}(0, \mathbb{I})$; 
        \STATE $\mathbf{v} = \nicefrac{\mathbf{v}}{||\mathbf{v}||_2}$;
        \STATE
         $\mathbf{v}_{next} = \mathbf{A}\mathbf{v}$;
    \ELSE
        \STATE $\mathbf{v}_{next} = \mathbf{A}\mathbf{v} - \beta_{m-1}\mathbf{v}_{prev}$;
    \ENDIF
    \STATE $\alpha_m = \mathbf{v}_{next}^T \mathbf{v}$;
    \STATE $\mathbf{v}_{next} = \mathbf{v}_{next} - \alpha_m \mathbf{v}$;
    \STATE $\beta_m = ||\mathbf{v}_{next}||_2$;
    \STATE $\mathbf{v}_{next} = \nicefrac{\mathbf{v}_{next}}{\beta_m}$;
    \STATE $\mathbf{v}_{prev} = \mathbf{v}$;
    \STATE $\mathbf{v} = \mathbf{v}_{prev}$;
\ENDFOR
\STATE \[\mathbf{T}_m =
\begin{bmatrix}
\alpha_1 & \beta_1  &        & \\
\beta_1  & \alpha_2 &        & \\
         &          & \ddots & \beta_{M-1} \\
         &          & \beta_{M-1} & \alpha_M \\
\end{bmatrix};\]
\STATE $\{\lambda_i\}_{i=1}^M, \{\mathbf{y}_i\}_{i=1}^M = \text{eig}(T_M)$;
\STATE $\{t_k\}_{k=1}^K= \text{linspace}(-1,1,K)$;
\FOR{$k = 1 \cdots K$}
\STATE $\sigma = \frac{2}{(M-1)\sqrt{8 \log(\kappa)}}$;
\STATE $\omega_k = \sum_{i=1}^M \mathbf{y}_i^i[1]^2 g_\sigma(t - \lambda_i) $;
\ENDFOR
\RETURN $\{\omega_k\}_{k=1}^K$
\end{algorithmic}
\end{algorithm}

\subsection{More Results:} 
We ran the layerwise eigenspectrum analysis further and present the results on ResNet-18 and DenseNet in Figs \ref{fig_spectrum_vgg11nb_cifar10}, \ref{fig:spectrum_den_cif10} when trained on the CIFAR10 dataset. Also, the layerwise t-SNE plots obtained upon clustering \G when a DenseNet is trained on CIFAR-10 is shown in Fig \ref{fig:den-cif10-tsne}.

\begin{figure}
    \centering
    \includegraphics[width=\linewidth]{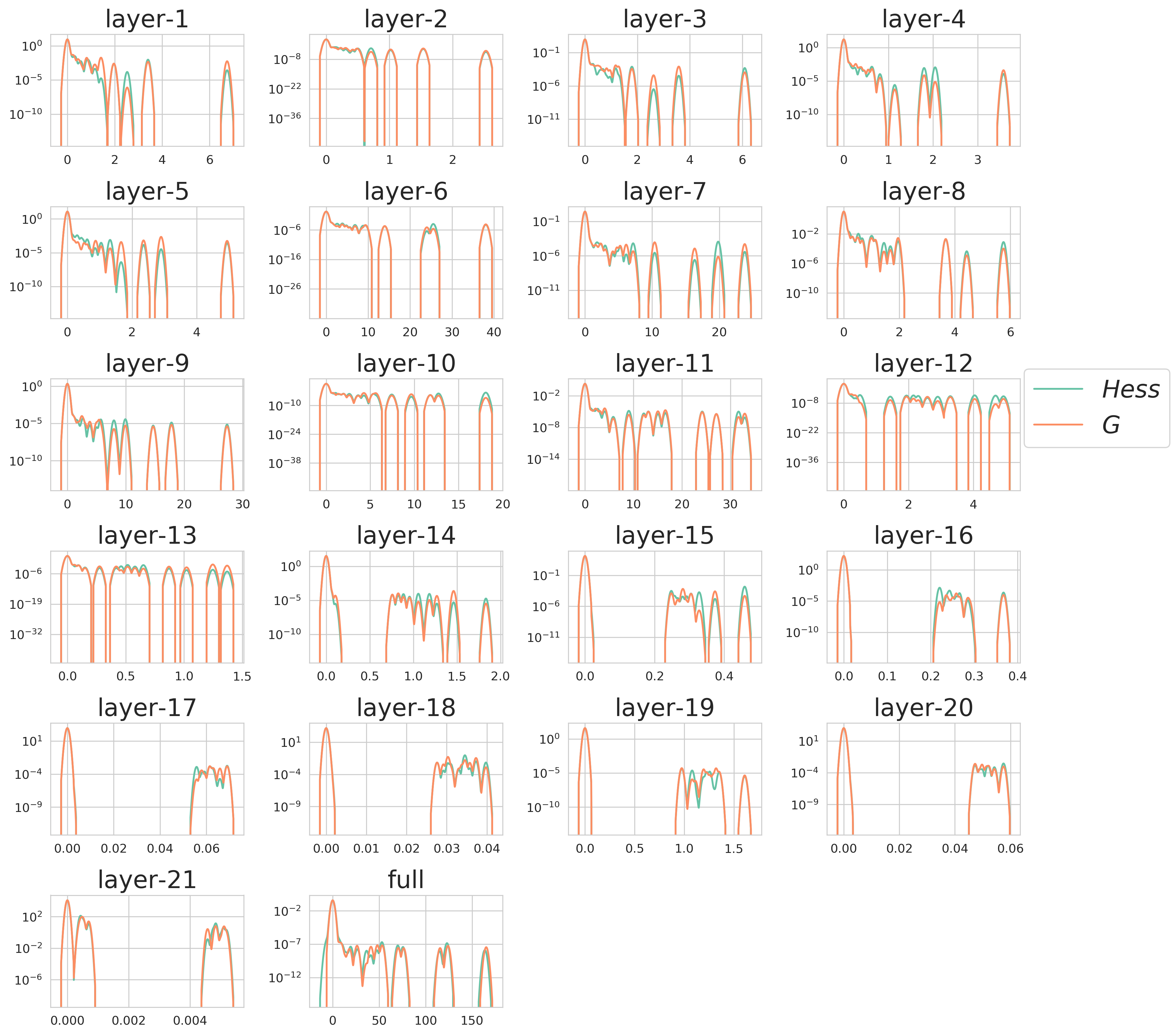}
    \caption{Eigenspectra of \HessL and \GL on  ResNet model trained on CIFAR-10 (Best viewed in color). Note that where the graph of the Hessian is not visible, the spectra of Hessian and G overlap completely. The last subplot ``full'' refers to eigenspectra of the \Hess and \G of the entire network;}  
    \label{fig_spectrum_vgg11nb_cifar10}
\end{figure}

\begin{figure}
    \centering
    \includegraphics[width=0.95\linewidth]{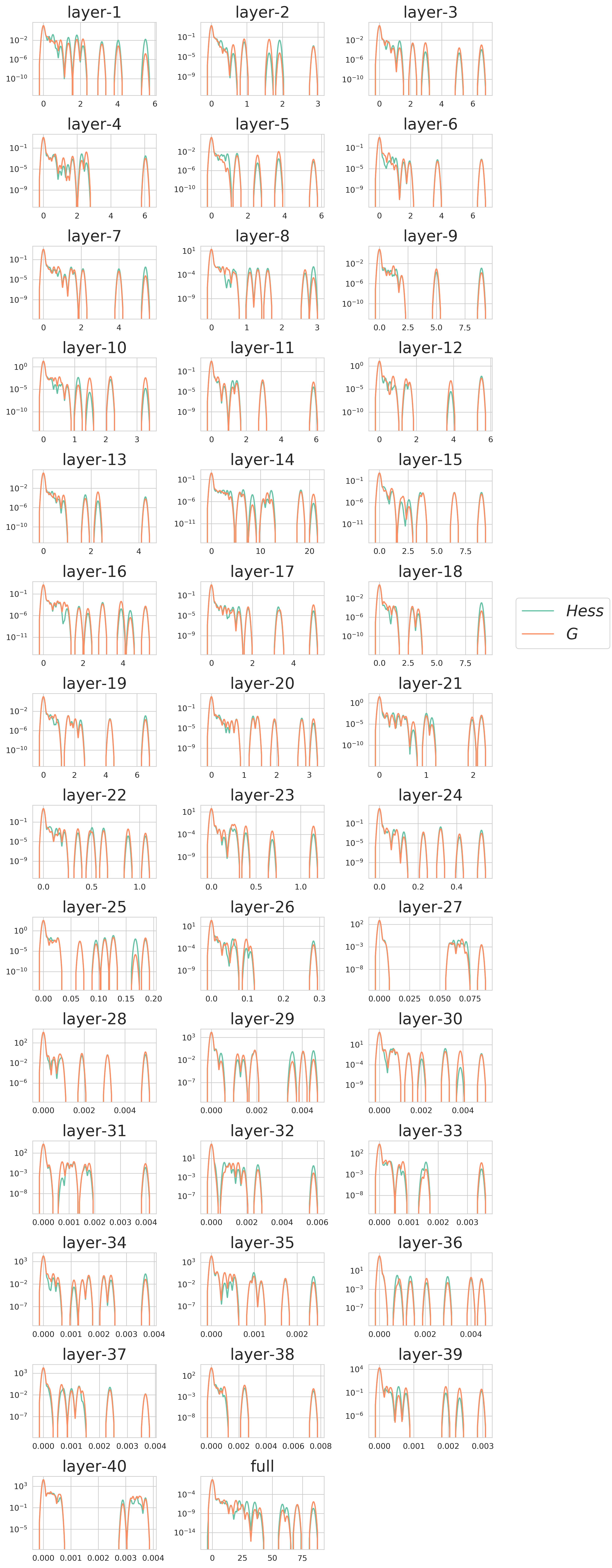}
    \caption{Eigenspectra of \HessL and \GL on DenseNet trained on CIFAR-10 (Best viewed in color)}
    \label{fig:spectrum_den_cif10}
\end{figure}

\begin{figure}
    \centering
    \includegraphics[width=\linewidth]{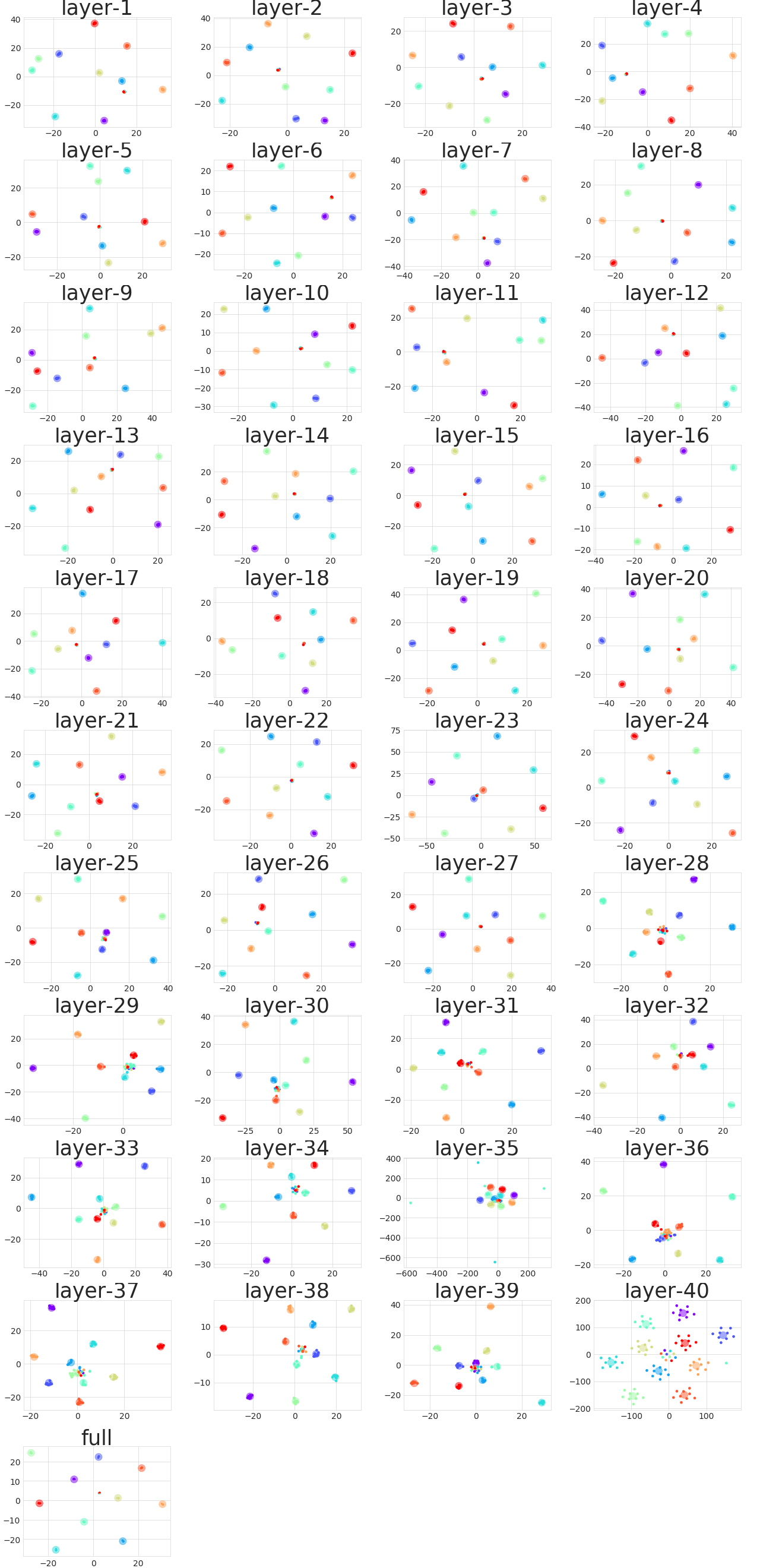}
    \caption{Layerwise two-dimensional t-SNE embeddings of $\delta_{c}$, obtained as a decomposition of \GL in \protect \cite{papyan2018full}  show $C$ clusters in each subplot (DenseNet on CIFAR10). The last subplot ``full'' corresponds to $\delta_{c}$ of the entire \G matrix.}
    \label{fig:den-cif10-tsne}
\end{figure}

\begin{figure}
  \begin{center}
    \includegraphics[scale=0.37]{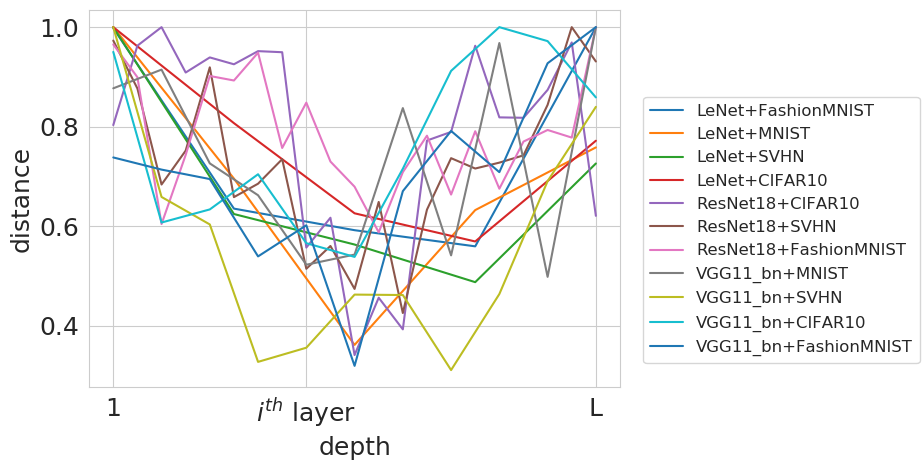}
    \caption{Plots of normalized Jensen–Shannon divergence between the spectra of $\mathbf{Hess}_l; \forall l = 1 \cdots L$ and $\mathbf{Hess}$ of different DNN models across datasets.}
     \label{fig:match-distros-kl}
  \end{center}
\end{figure}

Continuing our analysis further on studying the distance between the spectra of \HessL and \Hess, we calculated the Jensen–Shannon divergence between the layerwise Hessian and full network Hessian (instead of Wasserstein diatance used in the main paper). The results are shown in Fig \ref{fig:match-distros-kl}. The results corroborate the inferences made in the main paper. Even with a different distance metric, it can be observed that the spectra of \Hessl of middle layers of DNN models are very similar to spectrum of the overall \Hess. Such an observation has not been made hitherto, to the best of our knowledge. 

In continuation to Fig 4 in the main paper, we present results on the evolution of the layerwise Hessian eigenspectra using $\lambda_{max}$ (we presented the evolution of the full Hessian's $\lambda_{max}$ in Fig 4). Fig \ref{fig_trace_evolution_cifar10} shows how the $\lambda_{max}$ of each layerwise Hessian of ResNet-18 when trained on CIFAR-10 evolves as learning progresses. It can be observed that $\lambda_{max}$ of each layerwise Hessian goes down over training and also the middle layers have values of $\lambda_{max}$ closest to that of the entire network across the epochs.

\subsection{Ablation Studies and Analysis:} 
We conducted further experiments to study the performance of the proposed method more closely. As a first step, we studied the effect of the weighting co-efficient, $\gamma$  (Eqn \ref{eqn_htr_objective}), of the proposed HTR method. We varied $\gamma$ from $1e^{-8}$ to $1e^{-1}$ and studied the performance of HTR. Figure \ref{fig_gamma_ablation} shows the results, which shows that a weighting co-efficient around $1e^{-2}$ performed the best in this experiment. Lower values gave worse performance, showing the usefulness of the HTR term.

 

We also performed a wall-clock time comparison of training with and without HTR on various networks, and observed that the trace penalization takes approximately twice the time to finish across the different experiments. Table \ref{tab_wall_clock_times} reports these results. We can further accelerate the speed of HTR by penalizing the Hessian trace of specific layers in specific epochs of training, based on our understanding of the training process, or adaptively choose the weight co-efficients of the layers of the Hessian with a sparsity constraint to ensure that only few layers are considered for regularization in each iteration. This will make computations cheaper, and we leave this interesting direction for future work.
\begin{figure}[h]
  \begin{center}
    \includegraphics[scale=0.4]{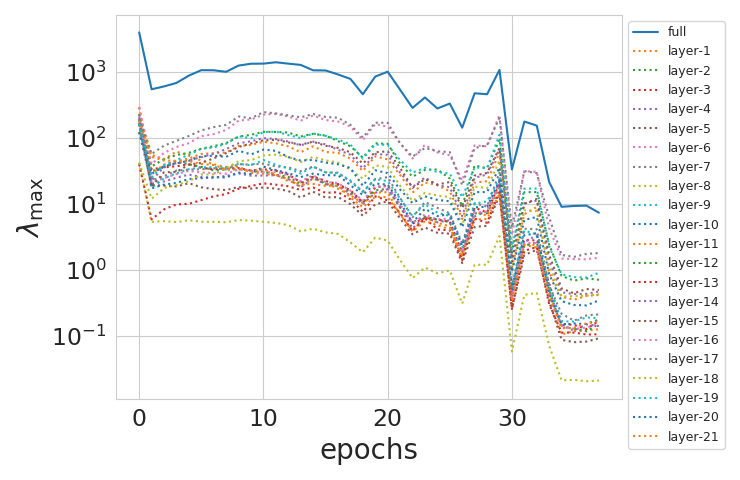}
    \caption{Evolution of $\lambda_{\max}$ of ResNet-18 when trained on CIFAR-10}
    \label{fig_trace_evolution_cifar10}
  \end{center}
\end{figure}
\begin{figure}[h]
   \begin{center}
     \includegraphics[width=0.85\linewidth]{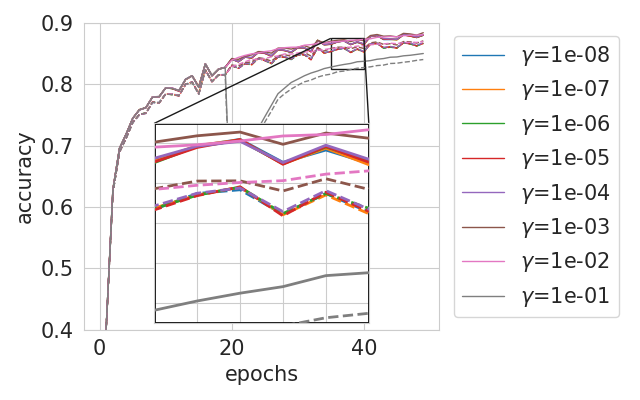}
     \caption{Ablation study on regularization parameter $\gamma$ in HTR, LeNet+FMNIST (Best viewed in color, solid lines denote train accuracy, dashed lines denote test accuracy)}
     \label{fig_gamma_ablation}
   \end{center}
 \end{figure}
\begin{table}[h]
\small
\footnotesize
    \centering
    \begin{tabular}{|c|c|c|}
        \hline \hline
        Model + Dataset &  freq & ratio\\
        \hline \hline
        \multirow{4}{*}{LeNet+MNIST} &  25 & 1.07 $\pm$ 0.13 \\
                                     &  50 & 1.14 $\pm$ 0.11 \\
                                     & 100 & 1.14 $\pm$ 0.15 \\
                                     & 200 & 1.00 $\pm$ 0.09 \\
                                     \hline
        \multirow{4}{*}{VGG11+CIFAR10} &  25 & 3.85 $\pm$ 0.06 \\
                                       &  50 & 2.40 $\pm$ 0.01 \\
                                       & 100 & 1.71 $\pm$ 0.01 \\
                                       & 200 & 1.42 $\pm$ 0.03 \\
                                       \hline
        \multirow{4}{*}{VGG11-bn+CIFAR10} &  25 & 4.93 $\pm$ 0.03 \\
                                          &  50 & 2.98 $\pm$ 0.02 \\
                                          & 100 & 1.99 $\pm$ 0.01 \\
                                          & 200 & 1.49 $\pm$ 0.01 \\
                                          \hline
        \multirow{4}{*}{ResNet18+CIFAR10} &  25 & 17.19 $\pm$ 0.06 \\
                                          &  50 &  9.09 $\pm$ 0.01 \\
                                          & 100 &  5.07 $\pm$ 0.05 \\
                                          & 200 &  3.00 $\pm$ 0.01 \\
                                          \hline
        \multirow{4}{*}{VGG11-bn+CIFAR100} &  25 & 3.01 $\pm$ 0.02 \\
                                           &  50 & 2.00 $\pm$ 0.01 \\
                                           & 100 & 1.50 $\pm$ 0.01 \\
                                           & 200 & 1.25 $\pm$ 0.00 \\
                                           
   \hline \hline
    \end{tabular}
    \caption{Ratios of wall clock times per epoch between proposed HTR regularization and Ordinary (SGD) training}
  \label{tab_wall_clock_times}
\end{table}

\paragraph{Connection between Hessian Trace and Eigenspectrum:}
When we penalize only the Hessian trace during training, there might exist situations when $\lambda_{min}$ of the Hessian can become negative (and potentially large in magnitude) making learning unstable. The Hessian trace can be near zero under three conditions: (i) All eigenvalues are near zero and positive; (ii) Eigen spectrum is symmetric about zero, or (iii) Eigen spectrum is asymmetric about zero, but with an equal density on the positive and negative side.  The above conditions, when combined with the fact that $\lambda_{\max}$ also goes to near zero (which our results in Sec 4 show) largely eliminates condition (iii). When all eigenvalues are positive and near zero, trace penalization leads to a flat minima, and is useful as has been shown in earlier work. When eigen distribution is symmetric around zero, trace penalization is still well- behaved as long as $\lambda_{\max}$ is reduced -  due to the symmetry assumption, the magnitude of $\lambda_{\min}$, even if negative, is also close to zero. When there is an equal amount of energy on both positive and negative sides of the eigen spectrum, the reduction of $\lambda_{max}$ as learning progresses ensures that the magnitude of $\lambda_{\min}$ must be close to zero. Thus, our empirical findings suggest that as trace and $\lambda_{\max}$ of the Hessian go to zero over training, a flat minimum ensues - thus motivating the HTR method.

\subsection{Connection to Neuroscience:} 
The observation of middle layers being similar to the overall network has been made in studies conducted in neuroscience too. The work by \cite{BEAULIEU1993284} studied how the neuronal density of parts of the brain, viz. occipital, parietal, and frontal, varies in a layerwise manner, and made similar conclusions. It was found that the neuron density in the middle layers of the brain is equal to the overall average neuronal density of the brain. This correlation between our observation of the middle layer Hessian spectrum closely matching the overall spectrum, and the middle layer neuronal density approximately equal to the average overall density is an interesting connection between the two fields.


\appendix

\end{document}